# Digit Recognition Using Convolution Neural Network

Kajol Gupta

Department of Computer Science & Engineering, Priyadarshini Institute of Engineering & Technology, Nagpur, India

Email: kajolgupta949@gmail.com

## Abstract

**In pattern recognition, digit recognition has always been a very challenging task. This paper aims to extracting a correct feature so that it can achieve better accuracy for recognition of digits. The applications of digit recognition such as in password, bank check process, etc. to recognize the valid user identification. Earlier, several researchers have used various different machine learning algorithms in pattern recognition i.e. KNN, SVM, RFC. The main objective of this work is to obtain highest accuracy 99.15% by using convolution neural network (CNN) to recognize the digit without doing too much pre-processing of dataset.**

## 1. Introduction

Digit recognition is an ability of a machine to take a digits as input and recognize which digit it is. It is very helpful in various applications such as in passwords to recognize the valid user. Digit image recognition plays an vital role in pattern recognition [1]. It has a wide range of applications in real life, such as in user identification, verification and handwritten digits recognition on bank check etc. Over the past few decades, various machine learning methods have been used for effective handwritten digit recognition [1][2][3][4], such as Linear and Non-Linear Classifier [2], Support Vector Machines (SVMs) [5] and many more etc. The performance of handwritten digit image recognition still needs to be improved due to the various reasons that offline recognition makes dynamic information (such as stroke and writing style) varies in size, stroke thickness, rotation, deformation [6][7][8], etc. In addition, to meet industry requirement, handwriting digit recognition systems must have highest accuracy and robustness to variations in handwriting style [8][9].

Handwritten character recognition is one of the practically important problems in pattern recognition applications. The main problem lies within the ability to produce an efficient algorithm that can recognize hand written digits and in which input is submitted by users through the scanner, tablet, and other digital devices. In the area of digit recognition system, identification of digit from where best discriminating features can be extracted is one of the important tasks. Various kinds of region sampling techniques are used in pattern recognition, to locate such regions [10]. Handwritten digits recognition is a well-researched subarea within the field that is concerned with learning models to distinguish pre-segmented handwritten digits. It is one of the major important issue in data mining, machine learning, pattern recognition along with many other disciplines of artificial intelligence [11].

In this paper, we are recognizing the digits from the images with better accuracy. Images is in the form of pixel values. To recognize the digits, we are using CNN model in which including various types of layers Convolution2D, MaxPooling2D, Dropout, Flatten, Dense.

## 2. Related Work

CNN is playing a vital role in many sectors like image processing. It has a powerful impact on various fields. In research work, it has shown that Deep Learning algorithm like multilayer CNN using Keras with Theano and Tensorflow which gives the highest accuracy in comparison with the other machine learning algorithms like SVM, KNN & RFC. Due to its highest accuracy, Convolutional Neural Network (CNN) is being used on a large scale in image classification, video analysis, etc [12]. Current algorithms are already pretty good at learning to recognize digits. It is really challenging for researchers to get better performance with low error. There are several research work, HyeranByun et al. [13] attempted the work using SVM algorithm on MINST dataset achieved 97.3% accuracy on Test data. Dan ClaudiuCiresan et al. [14] has proposed the work with Simple Neural network and back propagation used MNIST dataset 0.7% error rate i.e. 99.1% accuracy required higher processor, high cost, time consuming. Parveen Kumar, Nitin Sharma and Arun Rana [15] illustrated work to recognize a handwritten character using SVM classifier and MLP Neural Network. Different kernels are used like linear kernel, polynomial kernel, and quadratic kernel-based SVM classifiers. The linear kernel gives best accuracy of 94.8% among all the three kernels used. Retno Larasati et al. [16] that ensemble neural networks and ensemble decision tree that can classify MNIST and USPS dataset achieved accuracy upto 84%. T.Siva Ajay [17] has also proposed the work, in that it can be achieved by the use of convolutional neural networks the higher rate of accuracy in handwritten digit recognition task. The CNN implementation which is made easy with the use of LeNet engineering. As a result, the accuracy greater than 98% is obtained.

## 3. Dataset

Dataset is taken from Kaggle. The dataset contains total 42000 grey-scale images. Images is in the form of pixel values. Each image is 28 pixels in height and 28 pixels in width, 784 pixels in total. Each pixel has a single pixel-value associated with it, indicating the lightness or darkness of that pixel, higher pixel value indicates darker. This pixel-value is an integer between 0 and 255. The dataset contains 785 columns. The first column is label which is the digit that was drawn by the user. The rest of the columns contain the pixel-values of the associated image.

## 4. Normalization of Dataset

Normalization is used to scale the data of an attribute so that it falls in a smaller range, such as -1.0 to 1.0 or 0.0 to 1.0. It is generally useful for classification algorithms. The train and validation dataset contains 33600 and 8400 images respectively. To normalize, here we are dividing by a maximum pixel value to all pixel values. So they are normalized all the values on the same scale. The visualization of dataset in the form of image is represented in Figure 1.

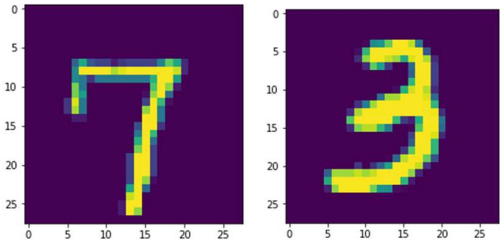

Fig. 1. Visualization of dataset in the form of image

## 5. Proposed Model Architecture

To recognize the digits, convolutional neural network with one input layer, three Convolution2D layers, two MaxPooling2D layers, three Dropout layers, one Flatten layer, two Dense layers has used. Demonstration of proposed Convolution Neural Network Model is illustrated in Figure 2.

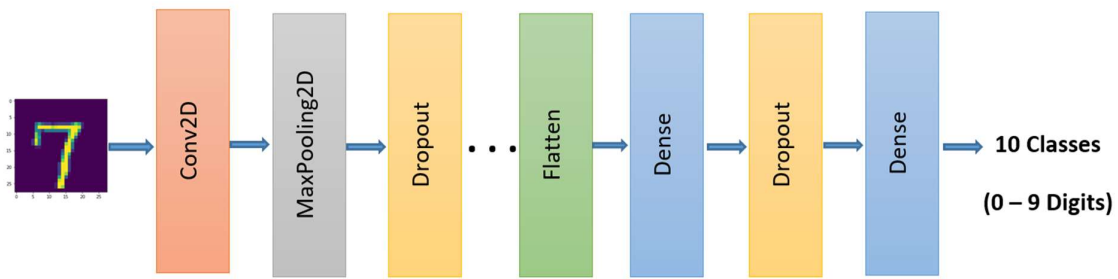

Fig. 2. Demonstration of proposed Convolution Neural Network Model

The input layer consists of 28 by 28 pixel images which means that the network contains 784 neurons as input data. The input pixels are grayscale with a value 0 for a white pixel and 1 for a black pixel. The convolution layer which is responsible for feature extraction from an input image. This layer performs convolution operation at small localized areas by convolving a filter with the previous layer. ReLU is used as an activation function at the end of each convolution layer. The MaxPooling layer which reduces the output information from the convolution layer and reduces the number of parameters. A Flatten layer is applied after the pooling layer which is responsible for conversion of the 2D featured map into a 1D feature vector. Dense layer is a classic fully connected layer. It is fully connected and connects every neuron from the previous layer to the next layer. Dropout is a regularization technique which aims to reduce the complexity of model and prevents from over-fitting. In this work, Dropout of 0.3 has used. Illustration of Flow of layers in proposed Convolution Neural Network for digit recognition in Table. 1.

| Layers | Output Shape |
|---|---|
| Input | (28, 28, 1) |
| Conv2D | (28, 28, 32) |
| MaxPooling2D | (14, 14, 32) |
| Dropout | (14, 14, 32) |
| Conv2D | (14, 14, 64) |
| MaxPooling2D | (7, 7, 64) |
| Dropout | (7, 7, 64) |
| Conv2D | (7, 7, 64) |
| Flatten | (3136) |
| Dense | (64) |
| Dropout | (64) |
| Dense | (10) |

**Table. 1. Flow of layers in proposed Convolution Neural Network for digit recognition**

## 6. Results

The accuracy and loss of the model has shown in the form of following graphs. To achieve a better accuracy, model required 15 number of epochs. The categorical cross entropy has used as loss function. Adam has used as an optimizer. The validation dataset contains 8400 images which has used to analysing the performance of model. The error rate of train and validation dataset is 02.63% and 02.93% respectively which is illustrated in Figure 3. Train and validation accuracy is 99.19% and 99.15% respectively which is illustrated in Figure 4. The accuracy and loss is also calculated in terms of confusion matrix.

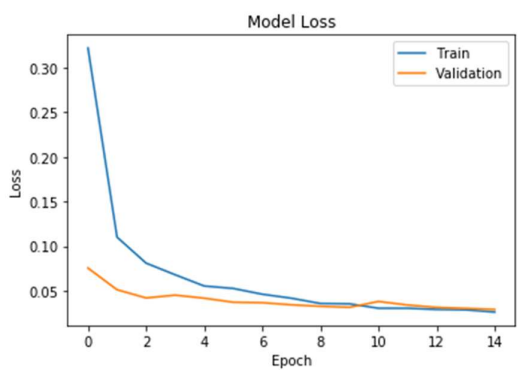
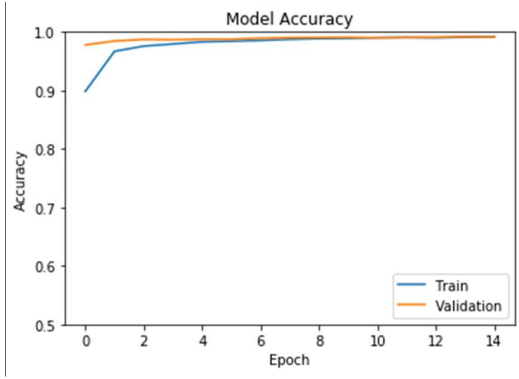

**Fig. 3. Loss Curve**              **Fig. 4. Accuracy Curve**

Confusion Matrix

A confusion matrix is a performance measurement technique for classification problems. It is a kind of table which helps you to know the performance of the classification model on a set of validation dataset for which the true values are known. Performance of such systems is commonly evaluated using the data in the matrix. The following table shows the confusion matrix for a ten class. The diagonal elements which representing correctly classified and other elements which representing misclassified by the model. The better accuracy has found in predicting 0 digit i.e. 812 out of 816 elements and the other which has correctly classified by the model for 1, 2, 3, 4, 5, 6, 7, 8, 9 are 904 out of 909, 840 out of 846, 930 out of 937, 830 out of 839, 689 out of 702, 781 out of 785, 885 out of 893, 826 out of 835, 832 out of 838 respectively. A confusion matrix of model prediction is represented in Table 2.

True Values

|   | 0 | 1 | 2 | 3 | 4 | 5 | 6 | 7 | 8 | 9 |
|---|---|---|---|---|---|---|---|---|---|---|
| 0 | 812 | 0 | 0 | 1 | 0 | 0 | 3 | 0 | 0 | 0 |
| 1 | 0 | 904 | 0 | 1 | 0 | 0 | 1 | 2 | 1 | 0 |
| 2 | 0 | 0 | 840 | 2 | 1 | 0 | 0 | 1 | 2 | 0 |
| 3 | 0 | 0 | 0 | 930 | 0 | 1 | 0 | 2 | 3 | 1 |
| 4 | 1 | 1 | 0 | 0 | 830 | 0 | 0 | 1 | 0 | 6 |
| 5 | 0 | 0 | 0 | 7 | 0 | 689 | 3 | 0 | 2 | 1 |
| 6 | 2 | 0 | 0 | 0 | 0 | 1 | 781 | 0 | 1 | 0 |
| 7 | 0 | 0 | 4 | 0 | 1 | 0 | 0 | 885 | 1 | 2 |
| 8 | 1 | 0 | 0 | 1 | 2 | 2 | 1 | 0 | 826 | 2 |
| 9 | 0 | 0 | 0 | 1 | 1 | 1 | 0 | 1 | 2 | 832 |

(Predicted Values on vertical axis)

**Table. 2. Confusion Matrix of model prediction**

## 7. Conclusion

The main objective of this paper is to develop a model which will recognize a digit with best accuracy. I have seen lot of papers in which they were used different machine learning algorithms i.e. KNN, SVM, RFC, etc. An implementation of Digit Recognition using CNN has been done in this paper. I have used convolution neural network model without requiring too much pre-processing techniques which achieved highest accuracy 99.15% with low error rate. Currently, implementation is done by using GPU.